\PassOptionsToPackage{table}{xcolor}
\documentclass{article} 
\usepackage{2026_conference,times}

\usepackage{amsmath,amsfonts,bm}

\def\eqref#1{equation~\ref{#1}}

\def\1{\bm{1}}

\DeclareMathAlphabet{\mathsfit}{\encodingdefault}{\sfdefault}{m}{sl}
\SetMathAlphabet{\mathsfit}{bold}{\encodingdefault}{\sfdefault}{bx}{n}

\usepackage{hyperref}
\usepackage{url}
\usepackage{booktabs}
\usepackage{multirow}
\usepackage{enumitem}
\usepackage{graphicx}
\usepackage{wrapfig}
\usepackage{makecell} 
\usepackage{amssymb}
\usepackage{wasysym}
\usepackage{comment}
\usepackage{subcaption}
\usepackage{booktabs,tabularx,makecell}
\usepackage{caption}
\usepackage{fontawesome}

\definecolor{iccvblue}{rgb}{0.21,0.49,0.74}
\hypersetup{
  colorlinks,
  citecolor=iccvblue,
  linkcolor=iccvblue,
  urlcolor=iccvblue}

\usepackage{booktabs}  
\usepackage{multirow}  
\usepackage{makecell}  
\usepackage{graphicx}  
\usepackage[table]{xcolor} 
\usepackage{multirow}
\definecolor{best}{HTML}{FFF2CC}  
\definecolor{second}{HTML}{E5F0E5} 
\definecolor{third}{HTML}{DCE6F2}   
\newcommand{\first}[1]{\cellcolor{best}#1}
\newcommand{\second}[1]{\cellcolor{second}#1}
\newcommand{\third}[1]{\cellcolor{third}#1}

\title{VideoWeave: Unlocking Geometric Consistency in Video Generation via Joint Geometry-Video Modeling}

\author{\textbf{Xunzhi Xiang}$^1$$^,$$^2$\textsuperscript{$*$}~~
    \textbf{Zixuan Duan}$^1$$^,$$^2$\textsuperscript{$*$}~~
    \textbf{Yabo Chen}$^2$\textsuperscript{$*$}~~
    \textbf{Zhengxuan Wei}$^1$~~
    \textbf{Guiyu Zhang}$^3$~~\\
    \textbf{Zixiao Gu}$^2$~~
    \textbf{Zhe Gao}$^1$~~
    \textbf{Haibin Huang}$^2$~~
    \textbf{Chi Zhang}$^2$~~
    \textbf{Qi Fan}$^1$\textsuperscript{$\dagger$}~~
    \textbf{Xuelong Li}$^2$\textsuperscript{$\dagger$}~~\\
    $^1$Nanjing University
    \quad
    $^2$Institute of Artificial Intelligence, China Telecom (TeleAI)\\
    $^3$Chinese University of Hong Kong, Shenzhen\\
}

\arxivfinalcopy 
\begin{document}

\maketitle

\begingroup
\renewcommand\thefootnote{}
\footnotetext{\textsuperscript{*}Equal contribution. \textsuperscript{$\dagger$} Corresponding authors.}
\endgroup


\begin{abstract}
Large-scale video diffusion models often fail to preserve 3D structure over time, causing geometric drift and implausible motion under viewpoint changes. Existing methods usually enforce geometric consistency by using explicit geometry reconstructions, such as depth maps, point clouds, or reconstructed 3D structures, to define conditions, supervision, or reward signals, making the generator sensitive to errors from upstream geometry pipelines. We propose VideoWeave, a latent-space post-training framework that uses implicit geometry-model features to constrain the generative distribution, providing a more flexible and non-rigid form of guidance that mitigates the impact of reconstruction errors from geometry models. Specifically, VideoWeave adapts these features into geometry latents and jointly models them with video latents in a shared denoising space, allowing geometry to shape the generative distribution during training. To support this process, we build GeoVid-80K, an 80K-video dataset with paired appearance and geometry representations. Experiments on text-to-video and image-to-video generation show that VideoWeave improves geometric coherence while preserving strong visual quality. \href{https://videoweave.github.io/}{VideoWeave project page}.
\end{abstract}


\section{Introduction}
As video generation moves toward world-level scene simulation \citep{sun2025worldplay, Voyager,chen2025teleworld,hy20,fantasyworld,gimworld}, 3D consistency becomes essential for applications such as autonomous driving \citep{gaia-1, drivedreamer} and embodied AI.
Current large-scale video diffusion models \citep{videocrafter, wan2025, kong2024hunyuanvideo, Cogvideo, xiang2025macro} can generate realistic frames, but they still struggle to maintain stable geometry over time \citep{moalign, mogan, chain, steerx}.
Since these models are trained mainly on 2D videos, they lack reliable 3D priors and often suffer from geometry drift under large viewpoint changes or long-duration generation.

The most straightforward remedy is to derive generation guidance from explicit geometric reconstructions, such as depth maps or point clouds, and use them as conditions \citep{uni3c, trajectorycrafter, World-consistent, gen3c}, supervision, or reward signals \citep{omnivdiff, geovideo, geometryascontext, vggrpo, wang2026world}. However, this approach typically introduces two critical bottlenecks. The first is \textit{error accumulation}: because these signals often rely on external off-the-shelf estimators \citep{cut3r, vggt, da3}, inaccuracies in depth or pose can be propagated into the generative process, which may compromise visual quality. The second is \textit{computational overhead}: when such signals are decoded, reconstructed, rendered, or repeatedly estimated to construct conditions, supervision targets, or rewards, the additional geometry pipeline can introduce non-negligible computation and scaling costs.

These limitations point to a more fundamental question: how should geometric information be incorporated into video generation without introducing the drawbacks of explicit reconstruction pipelines? Approaches that rely on decoded geometry to construct conditions, supervision, or rewards can expose the generator to errors and additional computational overhead from reconstruction or rendering pipelines. This motivates us to move geometric information away from explicit reconstructed outputs and instead place it inside the training-time generative state.

We address these bottlenecks with \textit{VideoWeave}. Instead of using explicit reconstruction results like point clouds or depth maps, which often propagate upstream errors, we extract internal features from geometry models to serve as a more flexible, non-rigid constraint. By coupling these implicit features with video latents in a shared diffusion space, joint distribution modeling naturally drives the video generation toward geometric consistency. VideoWeave integrates this prior progressively: it first adapts geometry features into the diffusion latent space without disrupting the original prior, then models them jointly with video as a unified denoising variable, and finally distills the joint score field into a compact student generator. The advantage of this design emerges during inference: the geometry latent is completely discarded. Consequently, VideoWeave generates 3D-consistent videos inherently, bypassing the computational overhead and rigid errors of explicit geometry estimation.

To support joint geometry-video learning, we further construct \textit{GeoVid-80K}, a curated large-scale dataset with paired appearance and implicit geometry representations. Unlike generic web video collections, GeoVid-80K is designed to emphasize rich depth cues, motion parallax, viewpoint changes, and temporally coherent scene structure. We mine videos from public spatial-aware datasets, licensed commercial footage, and open-access web sources, followed by automated filtering and expert review. The resulting dataset provides high-quality geometry-appearance pairs for learning 3D-aware video generative priors.

Compared with existing geometry-guided methods, VideoWeave changes the role of geometry from an explicit reconstruction-derived guidance signal into a temporary latent companion of the video during training. This positioning limits reliance on decoded geometry pipelines and lets the structural cue act through the denoising distribution itself. Experiments show that this design improves spatial coherence and long-range geometric stability while preserving visual fidelity and generation efficiency. In summary, our main contributions are as follows:
\begin{itemize}
    \item We propose VideoWeave, a latent-space post-training framework that treats geometry as a training-time latent variable shaping the video generative distribution, rather than as an explicit reconstruction-derived guidance signal.

    \item We introduce a progressive training strategy, including geometry latent adaptation, joint geometry-video diffusion modeling, and distribution-level score transfer, to bridge heterogeneous geometry and video representations while preserving the pretrained video prior.

    \item We construct GeoVid-80K, a curated dataset with rich depth cues, motion parallax, and viewpoint changes for learning implicit 3D priors.

\end{itemize}

\section{Related Work}

\noindent \textbf{Explicit Geometry Guidance.}
A direct way to improve 3D consistency in video generation is to introduce explicit geometric signals, such as depth maps \citep{geovideo, geometryascontext}, camera poses \citep{cameractrl, recammaster}, point clouds \citep{uni3c, gen3c}, or rendered 3D representations \citep{Voyager}. These signals can be used either as conditioning inputs during generation or as supervision/reward signals during training, providing direct structural constraints for scene layout and cross-view consistency. However, the effectiveness of such methods can be affected by the quality of the upstream geometry estimation or reconstruction pipeline. When the estimated geometry contains inaccurate depth, missing surfaces, pose drift, or reconstruction artifacts, these errors may be propagated into the generator, leading to distorted shapes, incorrect parallax, and inconsistent motion. Moreover, repeated 3D reconstruction, rendering, or geometry extraction for building conditions, supervision targets, or rewards can introduce substantial computational overhead, making these methods difficult to scale to modern large-scale video diffusion models. Therefore, while explicit geometry provides strong structural constraints, its robustness and scalability can be limited when the decoded geometry is noisy or expensive to obtain.

\noindent \textbf{Implicit Geometry Guidance.}
To reduce the dependence on decoded 3D reconstruction, some works leverage implicit features from pretrained geometry models to guide video generation \citep{geometryforcing, lavr, cinescene}. These features encode scene structure and spatial relationships in a continuous representation, making them more flexible than explicit depth maps or point clouds. However, geometry-model features and video generative latents are not naturally aligned. Forcing representation alignment between these heterogeneous spaces can disturb the pretrained generative prior and degrade visual quality.

\noindent \textbf{Joint Geometry-Video Modeling.}
Inspired by joint generative modeling~\citep{semanticgen}, recent works model geometry and video in a unified generative space, allowing structure and appearance to co-evolve during denoising. Several existing methods build joint objectives using decoded geometric targets or reconstruction-derived supervision/reward signals~\citep{geovideo, omnivdiff, mmphysvideo, videojam, gen3r}, making them sensitive to reconstruction quality. Other methods~\citep{dreamworld} compress geometry features with PCA and rely on additional loss scheduling. However, without representation adaptation, such task-agnostic linear projection cannot properly transform geometry-model features into generation-compatible latents, making it difficult to preserve generation-relevant structure and increasing optimization difficulty. In contrast, VideoWeave uses learnable geometry adaptation, progressive post-training, and score-level distillation to transfer 3D-aware priors more stably.

\section{Preliminary}
\label{sec:preliminary}

\noindent\textbf{Diffusion models.}
Diffusion models~\citep{ddpm, song-score, DiT} generate data by transforming Gaussian noise into samples from the target distribution, and have been widely used in image synthesis~\citep{ldm, dmd}, multi-view generation~\citep{nerf, cameractrl, 3dgs, recammaster}, and video generation~\citep{animatediff}.
Given clean data $\mathbf{x}$, the forward noising process is
\begin{equation}
    \mathbf{x}_t
    =
    F(\mathbf{x},\boldsymbol{\epsilon},t)
    =
    \alpha_t\mathbf{x}
    +
    \sigma_t\boldsymbol{\epsilon},
    \qquad
    \boldsymbol{\epsilon}\sim\mathcal{N}(\mathbf{0},\mathbf{I}),
    \label{eq:forward_process}
\end{equation}
where $\alpha_t$ and $\sigma_t$ control the signal-to-noise ratio.
A denoising model can be trained to predict the clean sample by
\begin{equation}
    \mathcal{L}_{\mathrm{denoise}}
    =
    \mathbb{E}_{\mathbf{x},t,\boldsymbol{\epsilon}}
    \left[
    \left\|
    \mathbf{x}
    -
    \hat{\mathbf{x}}_\theta(\mathbf{x}_t,t)
    \right\|_2^2
    \right].
    \label{eq:denoising_loss}
\end{equation}
Alternative parameterizations predict the noise $\boldsymbol{\epsilon}$ or velocity $\mathbf{v}$.
All parameterizations can be converted into a denoised estimate $\boldsymbol{\mu}(\mathbf{x}_t,t)$, giving the score estimate
\begin{equation}
    s(\mathbf{x}_t,t)
    =
    \nabla_{\mathbf{x}_t}\log p_t(\mathbf{x}_t)
    =
    -
    \frac{
    \mathbf{x}_t-\alpha_t\boldsymbol{\mu}(\mathbf{x}_t,t)
    }{
    \sigma_t^2
    }.
    \label{eq:score_from_denoised_estimate}
\end{equation}

\noindent\textbf{Distribution matching distillation.}
Distribution matching distillation (DMD)~\citep{dmd, SelfForcing} trains a few-step generator $G_\theta$ by matching its generated distribution to the real data distribution through score differences.
Given Gaussian input noise $\boldsymbol{\eta}$, the generator produces $\hat{\mathbf{x}}_0=G_\theta(\boldsymbol{\eta})$, which is perturbed as $\hat{\mathbf{x}}_t=F(\hat{\mathbf{x}}_0,\boldsymbol{\epsilon},t)$.
The generator is updated by
\begin{equation}
    \nabla_\theta\mathcal{L}_{\mathrm{DMD}}
    =
    -
    \mathbb{E}_{t,\boldsymbol{\eta},\boldsymbol{\epsilon}}
    \left[
    \Delta s(\hat{\mathbf{x}}_t,t)
    \frac{\partial G_\theta(\boldsymbol{\eta})}{\partial\theta}
    \right],
    \label{eq:dmd_gradient}
\end{equation}
where $\Delta s(\hat{\mathbf{x}}_t,t)=s_{\mathrm{real}}(\hat{\mathbf{x}}_t,t)-s_{\mathrm{fake}}(\hat{\mathbf{x}}_t,t)$.
Here $s_{\mathrm{real}}$ is provided by a large frozen teacher diffusion model, while $s_{\mathrm{fake}}$ is estimated by a small diffusion model trained on samples from $G_\theta$.

\noindent\textbf{Geometry foundation model.}
\label{sec:geometry_foundation_model}
Geometry foundation models encode spatial layouts and scene structures from input videos.
Given frames $\mathbf{I}\in\mathbb{R}^{B\times T_g\times H_I\times W_I\times3}$, a geometry encoder $\mathcal{E}_g$ extracts a multi-level token hierarchy $\{\mathbf{G}^{\ell}\}_{\ell=1}^{L}=\mathcal{E}_g(\mathbf{I})$.
Although these tokens can be decoded into explicit geometry such as dense depth, camera poses, or point clouds, we use them as continuous latent representations of scene structure.
This allows geometric knowledge to regularize video generation without exposing the generator to pixel-aligned reconstruction artifacts from decoded geometry.
We therefore bypass explicit geometry decoding and directly extract a dense geometry feature by
\begin{equation}
    \mathbf{G}
    =
    \Gamma\big(\{\mathbf{G}^{\ell}\}_{\ell=1}^{L}\big)
    \in
    \mathbb{R}^{B\times T_g\times C_G\times H_g\times W_g}.
    \label{eq:geometry_feature}
\end{equation}
In this paper, $\Gamma$ is treated as an experimental design parameter rather than a fixed heuristic choice, and its effect is analyzed in the ablation study.
\begin{figure*}[t]
    \centering
    \includegraphics[width=\textwidth]{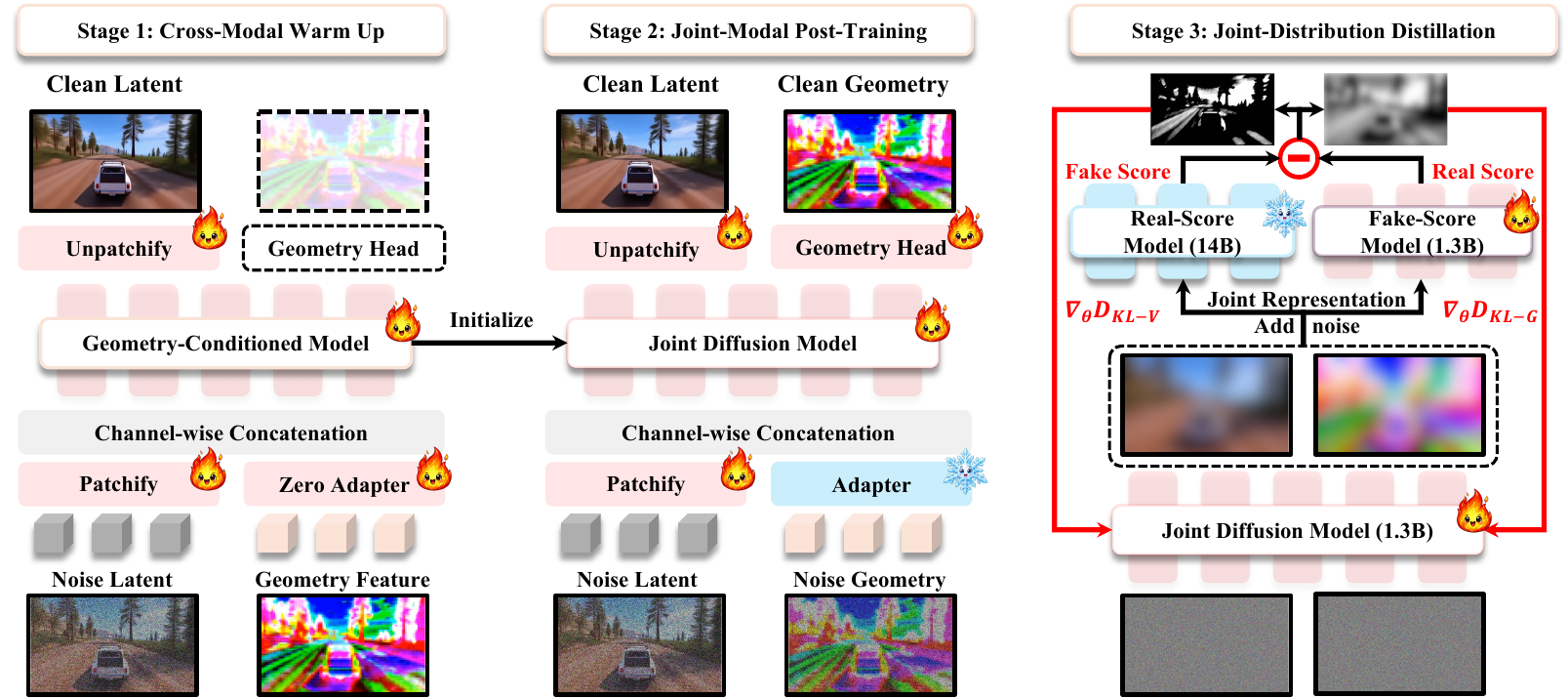}
    \caption{
    \textbf{Overall pipeline of VideoWeave.} The framework consists of three stages: (1) \textbf{Warm-up}: injecting adapted geometry features into the pretrained video prior; (2) \textbf{Joint Denoising}: learning a unified video--geometry diffusion space; and (3) \textbf{Distillation}: transferring the joint distribution to a compact generator for efficient and 3D-consistent video generation.
    }
    \label{fig:quality_diversity_training}
\end{figure*}

\section{Methodology}
\label{sec:method}

\subsection{Geometry Latent Adaptation}
\label{sec:geometry_adaptation}

Raw geometry features $\mathbf{G}$ and video latents $\mathbf{z}_0$ are defined on different feature grids, with mismatched temporal length, spatial resolution, and channel dimensionality.
The goal of this stage is to align their feature scales rather than their feature distributions.
Specifically, we introduce a lightweight geometry adapter $\mathcal{A}_\psi$ to convert $\mathbf{G}$ into a video-latent-compatible representation:
\begin{equation}
    \mathbf{g}_\psi
    =
    \mathcal{A}_\psi(\mathbf{G})
    \in
    \mathbb{R}^{B\times T\times C_g\times H\times W}.
    \label{eq:geometry_adapter}
\end{equation}
The adapter first matches the temporal and spatial dimensions of the video latent through uniform temporal subsampling and bilinear interpolation, and then projects the channel dimension with a trainable MLP.
After this adaptation, $\mathbf{g}_\psi$ is spatiotemporally aligned with $\mathbf{z}_0$ and channel-compatible for concatenation with the video latent.

This stage only resolves the scale mismatch between geometry features and video latents.
However, $\mathbf{g}_\psi$ is still produced from a geometry estimator and therefore follows a feature distribution different from the pretrained video latent space.
Directly using it for joint geometry-video modeling may introduce an abrupt distribution shift to the pretrained video diffusion model.
We therefore perform a geometry-latent warm-up stage to make the adapted geometry representation more compatible with the video generation prior.

\subsection{Geometry-Latent Warm-up}
\label{sec:teacher_warmup}

After scale alignment, we make $\mathbf{g}_\psi$ interpretable to the pretrained video diffusion model.
Instead of directly modeling video and geometry as a joint variable, we first introduce $\mathbf{g}_\psi$ through an auxiliary latent input branch while keeping the denoising target in the original video latent space.
Given $\mathbf{z}_t=F(\mathbf{z}_0,\boldsymbol{\epsilon}_z,t)$, we concatenate the noisy video latent and adapted geometry latent as $\tilde{\mathbf{z}}_t=[\mathbf{z}_t,\mathbf{g}_\psi]$.
To avoid disturbing the pretrained video prior at initialization, the original input projection $\mathbf{W}_z$ is expanded as $\tilde{\mathbf{W}}=[\mathbf{W}_z,\mathbf{0}_g]$, where $\mathbf{0}_g$ is a zero-initialized geometry projection.
The warm-up objective is
\begin{equation}
    \mathcal{L}_{\mathrm{warm}}
    =
    \mathbb{E}_{t,\mathbf{z}_0,\boldsymbol{\epsilon}_z}
    \left[
    \left\|
    v_{\theta}^{g}(\tilde{\mathbf{z}}_t,t,\mathbf{c})
    -
    \mathbf{u}_z
    \right\|_2^2
    \right],
    \label{eq:warmup_loss}
\end{equation}
where $\mathbf{u}_z=\boldsymbol{\epsilon}_z-\mathbf{z}_0$ is the target velocity for video denoising.

This warm-up stage optimizes the geometry adapter and the geometry input branch under the original video denoising objective.
As a result, $\mathbf{g}_\psi$ is not merely shape-compatible with $\mathbf{z}_0$, but also becomes usable by the pretrained video diffusion model as a soft structural cue.

\subsection{Geometry-Video Joint Diffusion Modeling}
\label{sec:joint_diffusion_modeling}

With the adapter optimized during warm-up ($\mathcal{A}_{\psi^\star}$), we obtain a geometry latent $\mathbf{g}_0=\mathcal{A}_{\psi^\star}(\mathbf{G})$ that is spatially and temporally aligned with the video latent $\mathbf{z}_0$. Instead of decoding geometry to build external conditions, supervision targets, or rewards, we treat the adapted geometry latent as part of a coupled training-time generative state $\mathbf{y}_0=[\mathbf{z}_0,\mathbf{g}_0]$, and train the model to capture the joint distribution $p(\mathbf{z}_0,\mathbf{g}_0)$ through a shared diffusion process.

A naive implementation would predict both video and geometry from the final output layer. However, this design can interfere with the pretrained video prior, since the last layers of video diffusion models are highly specialized for the denoising target in the video latent space and may suppress high-level 3D-related representations. In contrast, very early layers may not yet encode sufficient semantic and spatiotemporal structure. Therefore, we adopt a layer-aware joint modeling strategy: the video velocity is predicted by the original final output head, while the geometry velocity is predicted from an intermediate layer using a lightweight geometry head. This preserves the original video generation pathway while introducing geometry-aware learning at a representation level where 3D structure is still active.

Formally, given the noisy joint state $\mathbf{y}_t=F(\mathbf{y}_0,\boldsymbol{\epsilon}_y,t)$, we denote the hidden feature of the joint model at layer $\ell$ as
$\mathbf{h}^{\ell}_{\phi}(\mathbf{y}_t,t,\mathbf{c})$.
The joint velocity predictor $v_{\phi}^{j}$ consists of two components: the video velocity is predicted from the final layer, while the geometry velocity is predicted from the selected intermediate layer:
\begin{equation}
    v_{\phi,z}^{j}
    =
    \mathcal{O}_{z}\!\left(\mathbf{h}^{L}_{\phi}(\mathbf{y}_t,t,\mathbf{c})\right),
    \qquad
    v_{\phi,g}^{j,\ell}
    =
    \mathcal{O}_{g}\!\left(\mathbf{h}^{\ell}_{\phi}(\mathbf{y}_t,t,\mathbf{c})\right),
\end{equation}
where $\mathcal{O}_{z}$ is the original video prediction head, $\mathcal{O}_{g}$ is the geometry prediction head, $L$ denotes the final layer, and $\ell$ denotes the selected intermediate layer. The joint training objective is:
\begin{equation}
    \mathcal{L}_{\mathrm{joint}}
    =
    \mathbb{E}_{t,\mathbf{y}_0,\boldsymbol{\epsilon}_y}
    \left[
    \left\|
    v_{\phi,z}^{j}
    -
    \mathbf{u}_{z}
    \right\|_2^2
    +
    \lambda_g
    \left\|
    v_{\phi,g}^{j,\ell}
    -
    \mathbf{u}_{g}
    \right\|_2^2
    \right],
    \label{eq:joint_diffusion_loss}
\end{equation}
where $\mathbf{u}_{z}=\boldsymbol{\epsilon}_{z}-\mathbf{z}_0$ and $\mathbf{u}_{g}=\boldsymbol{\epsilon}_{g}-\mathbf{g}_0$ are the target velocities for video and geometry, respectively, and $\lambda_g$ balances the geometry learning objective.

This formulation changes the learning target from a conditional mapping $p(\mathbf{z}_0|\mathbf{g}_0)$ to a coupled geometry-video generative process. The video branch preserves the pretrained model's video-latent denoising pathway, while the intermediate geometry branch encourages the shared backbone to encode geometry-aware spatiotemporal dynamics. Because both branches share the denoising backbone, the geometry target regularizes intermediate representations used by the video branch. This shared denoising objective improves multi-view consistency and reduces shape drifting while avoiding decoded reconstruction or rendering signals as conditions, supervision targets, or rewards.

\subsection{Joint-Distribution Distillation}
\label{sec:joint_distribution_distillation}

The joint diffusion model improves geometry-video consistency but remains costly due to iterative denoising.
We therefore distill its joint distribution into a few-step generator $G_\omega$.
Given Gaussian noise $\boldsymbol{\eta}$ and condition $\mathbf{c}$, the student directly predicts $\hat{\mathbf{y}}_0=G_\omega(\boldsymbol{\eta},\mathbf{c})=[\hat{\mathbf{z}}_0,\hat{\mathbf{g}}_0]$, where $\hat{\mathbf{z}}_0$ and $\hat{\mathbf{g}}_0$ are the generated video and geometry latents.
The generated joint latent is perturbed as $\hat{\mathbf{y}}_t=F(\hat{\mathbf{y}}_0,\boldsymbol{\epsilon}_y,t)$.
Following DMD, the student is updated by
\begin{equation}
    \nabla_\omega\mathcal{L}_{\mathrm{JDD}}
    =
    -
    \mathbb{E}_{t,\boldsymbol{\eta},\boldsymbol{\epsilon}_y}
    \left[
    w(t)
    \Delta s^j(\hat{\mathbf{y}}_t,t,\mathbf{c})
    \frac{\partial G_\omega(\boldsymbol{\eta},\mathbf{c})}{\partial\omega}
    \right],
    \label{eq:jdd_gradient}
\end{equation}
where $\Delta s^j(\hat{\mathbf{y}}_t,t,\mathbf{c})=s_{\mathrm{T}}^j(\hat{\mathbf{y}}_t,t,\mathbf{c})-s_{\mathrm{G}}^j(\hat{\mathbf{y}}_t,t,\mathbf{c})$.
Here $s_{\mathrm{T}}^j$ is estimated by the frozen geometry-video joint teacher, while $s_{\mathrm{G}}^j$ is estimated by a fake-score model trained on samples from $G_\omega$.
This objective transfers the teacher's joint score field to the student, allowing the compact generator to preserve both visual fidelity and geometry-video consistency.

During inference, $G_\omega$ outputs $\hat{y}_0=[\hat{z}_0,\hat{g}_0]$, but only the video latent $\hat{z}_0$ is decoded by the video VAE, while $\hat{g}_0$ is discarded. Therefore, the geometry latent is not an inference-time input condition, but a training-time auxiliary variable used to shape the joint score field. After distillation, the generated video latent itself carries the distilled geometry-aware prior. VideoWeave thus improves 3D consistency without running a geometry encoder, reconstruction module, or renderer during inference, making it suitable for efficient video generation.

\section{GeoVid-80K Dataset}
We introduce \textbf{GeoVid-80K}, a large-scale dataset for world-consistent video generation with paired appearance and geometry signals. It is curated from unconstrained web videos and public video datasets by mining sequences with rich depth cues, motion parallax, and viewpoint changes, followed by temporally consistent geometry extraction to support joint geometry--appearance learning. 

\subsection{Data Collection}
We construct a high-resolution video dataset centered on pronounced camera motion, such as panning, tracking, and fly-throughs, to ensure the availability of dense multi-view spatio-temporal cues, which are essential for 3D prior learning. Our dataset encompasses both static scenes to capture pure background geometry and dynamic scenes featuring object interactions. Both categories span indoor and outdoor environments, further supplemented by aerial cinematography to capture large-scale, unbounded spatial structures.

\begin{figure*}[t]
    \centering
    \includegraphics[width=\textwidth]{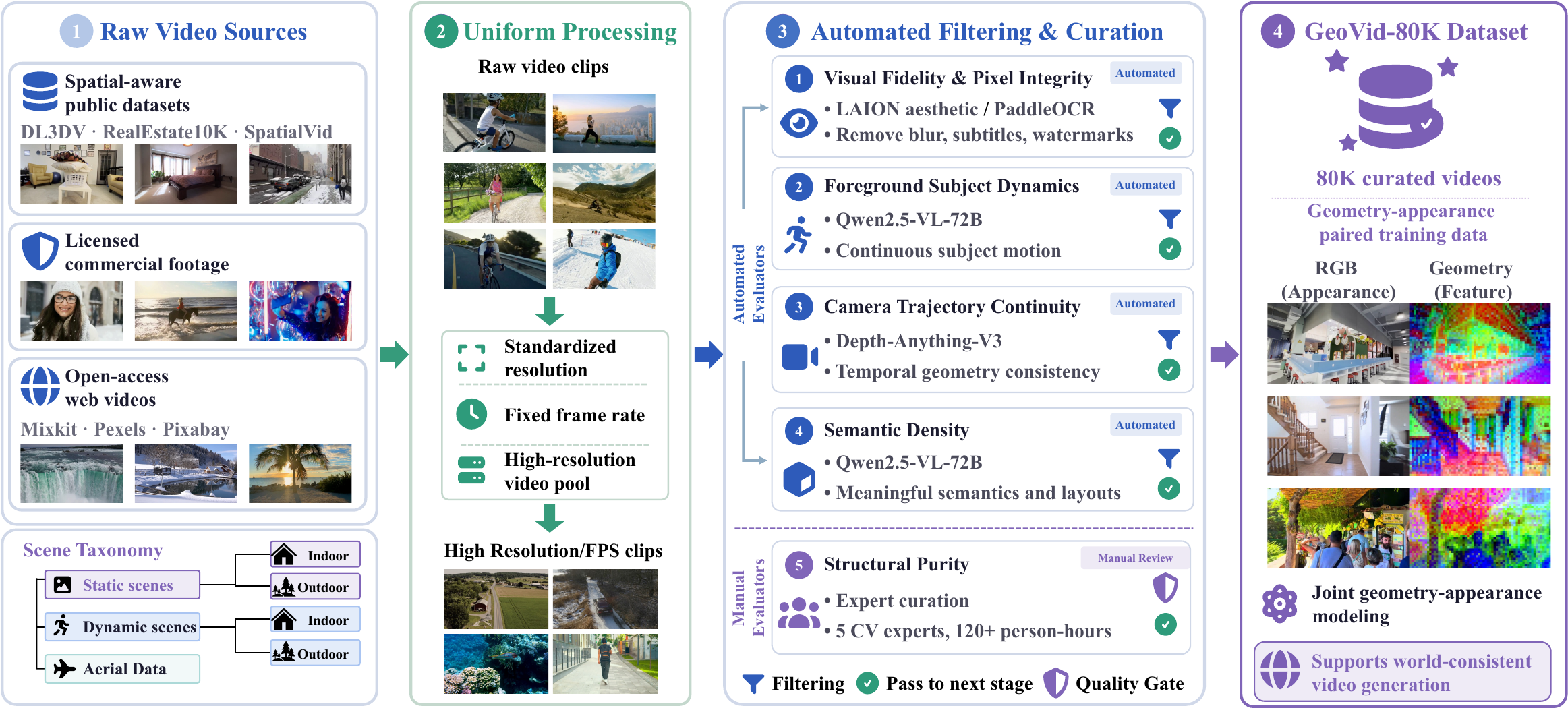}
    \caption{
    \textbf{Overview of the GeoVid-80K data curation pipeline.}
    Diverse raw videos are normalized, automatically filtered for visual, motion, geometric, and semantic quality, and finally reviewed by experts to obtain 80K geometry--appearance paired videos for world-consistent video generation.
    }
    \label{fig:data_pipeline}
\end{figure*}

To build this dataset, we aggregate videos from spatial-aware public benchmarks, specifically DL3DV \citep{ling2024dl3dv}, RealEstate10K \citep{realstate}, MiraData \citep{ju2024miradata}, and SpatialVid \citep{spatialvid}. We also collect licensed commercial footage containing professional motion trajectories along with high-quality web videos sourced from open-access platforms like Mixkit, Pexels, and Pixabay. All raw sequences are then unified to a standardized resolution and frame rate.

\subsection{Data Filtering and Curation}
\label{sec:data_filtering}

To ensure reliable geometry-video learning, we build an automated filtering pipeline followed by expert review. 
The pipeline filters videos from five aspects:

\noindent\textbf{Visual Quality.}
We use the LAION aesthetic predictor~\citep{laion} to filter out visually degraded clips, including blur, overexposure, severe compression artifacts, and poor composition.
PaddleOCR is further applied to remove videos with text overlays, logos or watermarks.

\noindent\textbf{Subject Motion.}
We use Qwen2.5-VL-72B~\citep{qwenvl} to identify clips with clear, continuous, and semantically meaningful foreground motion.
Videos with nearly static subjects, ambiguous dynamics, or artifact-driven motion are discarded.

\noindent\textbf{Camera Motion.}
We use Depth-Anything-V3~\citep{da3} to estimate temporal and geometric consistency.
This step retains clips with smooth viewpoint changes and removes videos with abrupt cuts, unstable camera motion, or inconsistent scene geometry.

\noindent\textbf{Semantic Content.}
We use Qwen2.5-VL-72B~\citep{qwenvl} to assess scene composition, object clarity, and semantic richness.
Clips with abstract content, unclear spatial layout, weak scene structure, or ambiguous main subjects are filtered out.

\noindent\textbf{Expert Review.}
Finally, domain experts conduct over 120 person-hours of manual review to remove remaining failure cases, including subtle geometric artifacts, unnatural motion, temporal discontinuities, and VLM misjudgments.

To facilitate future research and support reproducible evaluation, we plan to release GeoVid-80K together with the data processing and filtering pipeline after the necessary licensing and release checks.

\section{Experiments}

\noindent\textbf{Baselines.}
We evaluate VideoWeave under both text-to-video (T2V) and image-to-video (I2V) settings.
For T2V, we compare with OmniVDiff~\citep{omnivdiff}, GeoVideo~\citep{geovideo}, VideoGPA~\citep{videogpa}, and a same-budget Wan-SFT baseline.
For I2V, we compare with GeometryForcing~\citep{geometryforcing}, ViewCrafter~\citep{viewcrafter}, and Gen3C~\citep{gen3c}, which represent recent geometry-aware and camera-controllable image-to-video generation methods.
Specifically, Wan-SFT uses the same Wan initialization, GeoVid-80K data, and training budget as VideoWeave, but removes all geometry-aware components, isolating the effect of our training pipeline from dataset-only gains.

\noindent\textbf{Implementation Details.}
We use DA3~\citep{da3} to extract geometry features, and adopt Wan-14B and Wan-1.3B~\citep{wan2025} as the teacher and student generators, respectively.
Unless otherwise specified, $\Gamma$ concatenates layer-20 and layer-40 features from the DA3-Giant encoder before geometry-latent adaptation.
The geometry adapter and geometry prediction head are lightweight two-layer MLPs, and the geometry loss is applied at DiT transformer layer 16 with $\lambda_g=1.0$.
For I2V, the target camera trajectory is encoded with a dense Pl\"ucker ray embedding, concatenated with the noisy video latent along the channel dimension, and projected by the input layer together with the reference-image condition.
For T2V training, we use the full GeoVid-80K training set to learn geometry-aware video priors from diverse scenes and motions.
For I2V training, to ensure a fair comparison with camera-conditioned I2V baselines, we train only on the RealEstate10K training split rather than using the full GeoVid-80K dataset.
Training follows three stages: 2K steps for geometry-latent warm-up, 15K/20K steps for joint geometry-video training on Wan-14B/Wan-1.3B, and 1K steps for joint-distribution score distillation.
All stages are optimized with AdamW using a global batch size of 48.
The final distilled student is sampled with five inference steps.

\noindent\textbf{Evaluation Protocol.}
We evaluate VideoWeave under both text-to-video (T2V) and image-to-video (I2V) settings.
For T2V evaluation, we construct a test set of 100 prompts, covering indoor scenes, outdoor scenes, object-centric motion, human activities, and camera-motion scenarios.
For I2V evaluation, we construct a held-out test set by sampling 100 RealEstate10K videos that are excluded from the I2V training split.
We use the first frame of each video as the reference image and the corresponding camera trajectory as the target motion.
For each baseline, we use the publicly released checkpoint and follow its recommended inference configuration, including its adapted resolution, frame number, sampling scheduler, and guidance scale.
All methods are evaluated with the same fixed random seed to ensure consistent sampling conditions.
This protocol avoids re-tuning baselines for our setting and evaluates each method under its intended operating regime.
For metric computation, generated videos are converted to a unified format before evaluation.
We assess generated videos from three perspectives.

\noindent \textit{(1) General Video Quality}: we use VBench~\citep{Vbench} to evaluate background consistency, subject consistency, motion smoothness, imaging quality, and aesthetic quality.

\noindent \textit{(2) 3D Reconstruction Consistency}: we reconstruct each generated video with 3DGS, re-render it along the estimated camera trajectory, and compute PSNR~\citep{PSNR}, SSIM~\citep{SSIM}, LPIPS~\citep{LPIPS}, and MSE against the generated frames.

\noindent \textit{(3) Epipolar Consistency}: we use SIFT-based epipolar matching~\citep{epipolar} to measure cross-frame geometric stability, reporting Epipolar Error and Inlier Rate.
Since our focus is camera-induced 3D consistency rather than generic text-to-video benchmarking, we explicitly evaluate T2V models under prompts that require noticeable camera motion.
This avoids the trivial case where nearly static videos obtain inflated consistency scores simply because there is little viewpoint change.
We therefore augment all T2V prompts with ``noticeable camera movement, dynamic shot'' and retain sufficiently dynamic samples using VBench Dynamic Degree.

\begin{table*}[t]
\centering
\caption{
    \textbf{Quantitative Comparison on Text-to-Video Quality, 3D Geometry, and Consistency.}
    \colorbox{best}{\phantom{x}} indicates the best performance, \colorbox{second}{\phantom{x}} the second best, and \colorbox{third}{\phantom{x}} the third best.
}
\label{tab:main_results}
\resizebox{\textwidth}{!}{
\begin{tabular}{l | ccccc | cccc | cc}
\toprule
\multirow{2}{*}{\textbf{Method}} &
\multicolumn{5}{c|}{\textbf{VBench Metrics}} &
\multicolumn{4}{c|}{\textbf{3D Reconstruction Error}} &
\multicolumn{2}{c}{\textbf{3D Consistency Metrics}} \\
\cmidrule(lr){2-6} \cmidrule(lr){7-10} \cmidrule(l){11-12}
&
\makecell{\textbf{Background} \\ \textbf{Consistency} $\uparrow$} &
\makecell{\textbf{Subject} \\ \textbf{Consistency} $\uparrow$} &
\makecell{\textbf{Motion} \\ \textbf{Smoothness} $\uparrow$} &
\makecell{\textbf{Imaging} \\ \textbf{Quality} $\uparrow$} &
\makecell{\textbf{Aesthetic} \\ \textbf{Quality} $\uparrow$} &
\makecell{\textbf{PSNR} $\uparrow$} &
\makecell{\textbf{SSIM} $\uparrow$} &
\makecell{\textbf{LPIPS} $\downarrow$} &
\makecell{\textbf{MSE} $\downarrow$} &
\makecell{\textbf{Epipolar} \\ \textbf{Error} $\downarrow$} &
\makecell{\textbf{Inlier} \\ \textbf{Rate} $\uparrow$} \\
\midrule
OmniVDiff
& \first{0.9309} & \second{0.9230} & \second{0.9875} & 0.6135 & 0.5330 
& \third{19.5630} & \third{0.6157} & 0.4169 & \third{856.9066} 
& \second{9.2397} & \second{0.6133} \\

GeoVideo    
& \third{0.9182} & 0.8839 & 0.9646 & \third{0.6334} & 0.5340 
& 18.9986 & \second{0.6349} & \third{0.3975} & 1072.6943 
& 33.7536 & 0.3033 \\

VideoGPA 
& 0.8972 & 0.8155 & 0.9647 & 0.6314 & \second{0.5740} 
& 17.0392 & 0.5353 & 0.4061 & 1865.7591 
& \third{11.1430} & \third{0.5866} \\

Wan-SFT   
& 0.9086 & \third{0.9000} & \third{0.9712} & \second{0.6659} & \third{0.5633} 
& \second{19.7444} & 0.5913 & \second{0.3790} & \second{815.3085} 
& 19.9983 & 0.3935 \\

\midrule
\textbf{VideoWeave-T2V} 
& \second{0.9305} 
& \first{0.9530} 
& \first{0.9910} 
& \first{0.7383} 
& \first{0.5988} 
& \first{21.4994} 
& \first{0.7082} 
& \first{0.2987} 
& \first{500.8694} 
& \first{7.2780} 
& \first{0.7068} \\
\bottomrule
\end{tabular}
}
\end{table*}

\begin{figure*}[t]
    \centering
    \includegraphics[width=\textwidth]{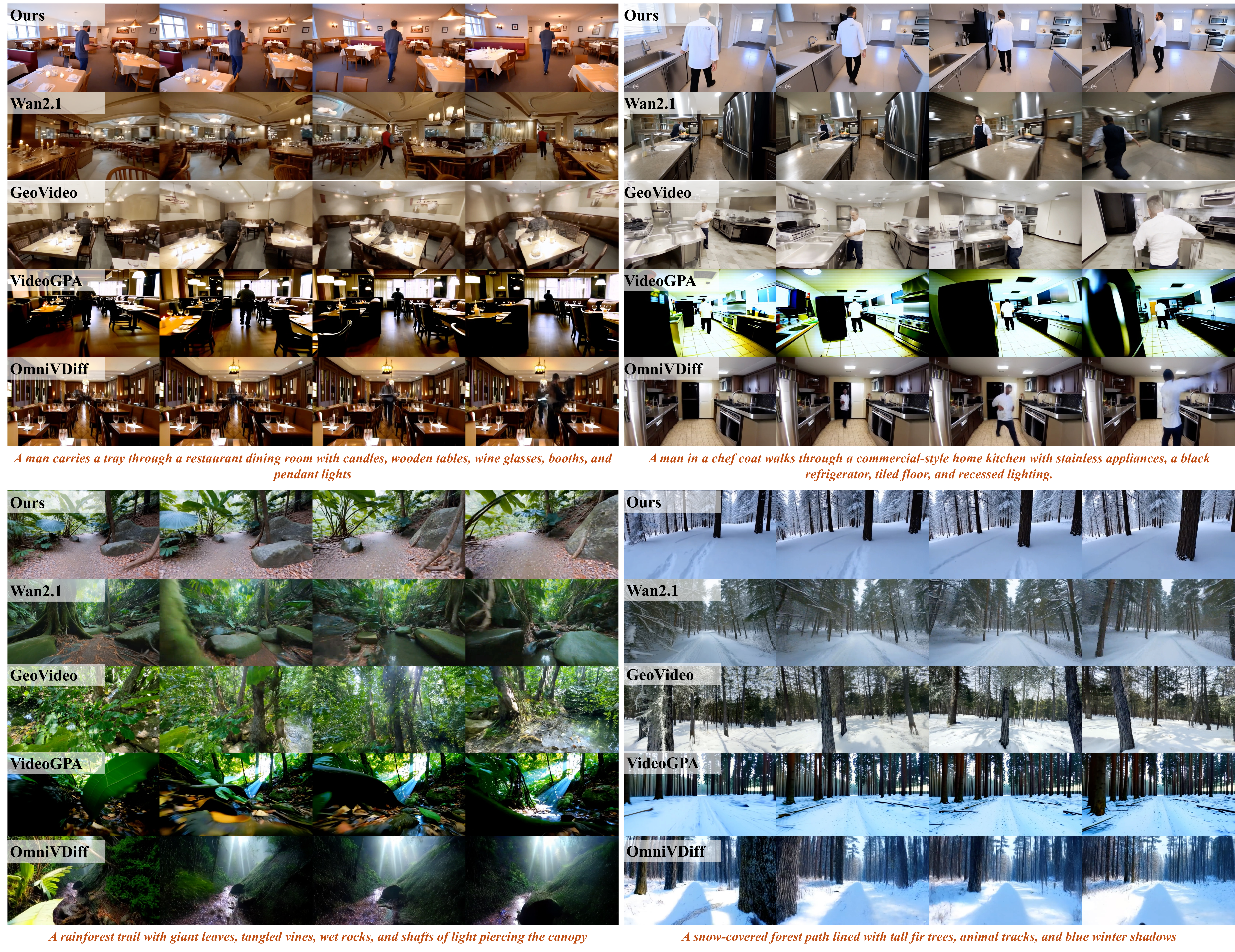}
\caption{
    \textbf{Qualitative Comparison of Text-to-Video Generation.} 
    VideoWeave produces more stable layouts and fewer geometric distortions on diverse indoor and outdoor scenes.
}
    \label{fig:qualitative_results}
\end{figure*}

\begin{figure}[t]
    \centering
    \includegraphics[width=\textwidth]{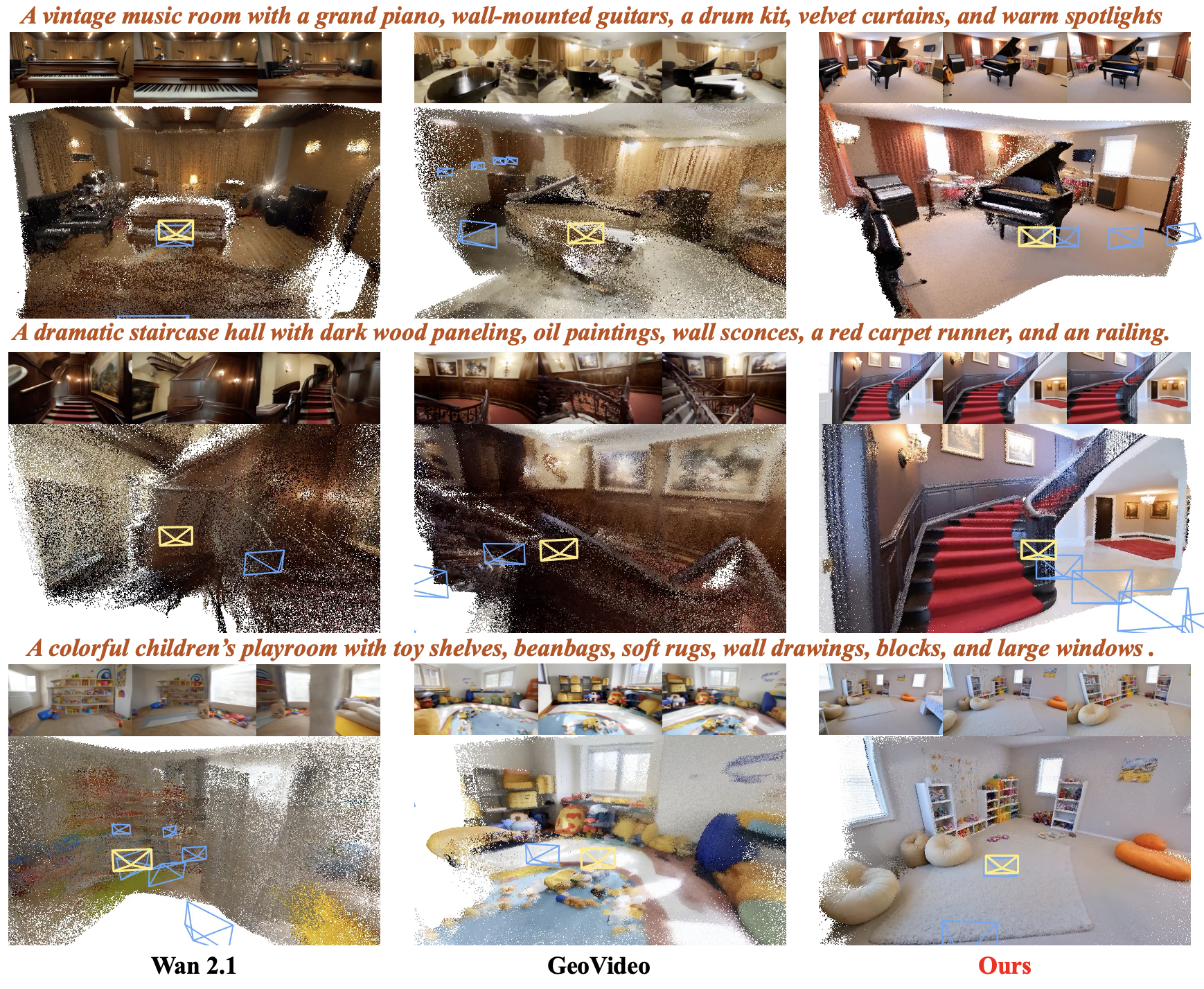}
    \caption{
\textbf{Qualitative Comparison of Reconstructed Point Clouds.}
Point clouds reconstructed from VideoWeave videos show more complete structures and better cross-view alignment.}
      \label{fig:compare-3d}
\end{figure}

\subsection{Main Results}
\noindent \textbf{Text-to-Video Quantitative Results.}
As shown in Table~\ref{tab:main_results}, VideoWeave achieves strong performance on both video quality and 3D consistency. On VBench, it obtains the best subject consistency, motion smoothness, and imaging quality, showing that geometry-aware modeling can improve geometric coherence while maintaining competitive visual generation quality. More importantly, VideoWeave consistently improves 3D reconstruction and epipolar consistency, achieving the best PSNR, SSIM, LPIPS, MSE, and epipolar error among all compared methods. These results show that VideoWeave improves not only the appearance of generated videos, but also their underlying geometric stability. Compared with pure 2D video generators and methods using external geometry supervision, VideoWeave learns an implicit geometry--video joint prior that is better aligned with the generation process, leading to more stable 3D structure over time.

\noindent \textbf{Text-to-Video Qualitative Results.}
We visualize the generated videos and their reconstructed point clouds in Fig.~\ref{fig:qualitative_results} and Fig.~\ref{fig:compare-3d}. VideoWeave produces more temporally stable videos with consistent scene layout, object geometry, and camera motion. In contrast, baseline methods often show geometry deformation, identity drift, and inconsistent spatial structure across frames. The point cloud results further highlight this difference. In our evaluated cases, VideoWeave yields more complete and better-aligned 3D reconstructions, while pure 2D video generators often produce more fragmented point clouds due to weak 3D consistency. Explicit 3D-supervised methods introduce geometric constraints, but their point clouds still contain noticeable noise and misalignment. These results show that implicit geometry-video joint modeling better aligns 3D structure with the video generation process, improving both visual temporal consistency and underlying 3D coherence.

\begin{table*}[t]
\centering
\caption{
    \textbf{Quantitative Comparison on Image-to-Video Quality, 3D Geometry, and Consistency.}
    \colorbox{best}{\phantom{x}} indicates the best performance, \colorbox{second}{\phantom{x}} the second best, and \colorbox{third}{\phantom{x}} the third best.
}
\label{tab:main_results_i2v}
\resizebox{\textwidth}{!}{
\begin{tabular}{l | ccccc | cccc | cc}
\toprule
\multirow{2}{*}{\textbf{Method}} &
\multicolumn{5}{c|}{\textbf{VBench Metrics}} &
\multicolumn{4}{c|}{\textbf{3D Reconstruction Error}} &
\multicolumn{2}{c}{\textbf{3D Consistency Metrics}} \\
\cmidrule(lr){2-6} \cmidrule(lr){7-10} \cmidrule(l){11-12}
&
\makecell{\textbf{Background} \\ \textbf{Consistency} $\uparrow$} &
\makecell{\textbf{Subject} \\ \textbf{Consistency} $\uparrow$} &
\makecell{\textbf{Motion} \\ \textbf{Smoothness} $\uparrow$} &
\makecell{\textbf{Imaging} \\ \textbf{Quality} $\uparrow$} &
\makecell{\textbf{Aesthetic} \\ \textbf{Quality} $\uparrow$} &
\makecell{\textbf{PSNR} $\uparrow$} &
\makecell{\textbf{SSIM} $\uparrow$} &
\makecell{\textbf{LPIPS} $\downarrow$} &
\makecell{\textbf{MSE} $\downarrow$} &
\makecell{\textbf{Epipolar} \\ \textbf{Error} $\downarrow$} &
\makecell{\textbf{Inlier} \\ \textbf{Rate} $\uparrow$} \\
\midrule
GeometryForcing         
& \first{0.9747} & \third{0.9653} & 0.9870 & \third{0.6918} & 0.4875 
& 17.9446 & 0.6417 & 0.4403 & 1246.8988
& - & - \\

ViewCrafter
& \second{0.9717} & 0.9621 & \third{0.9879} & 0.6898 & \second{0.5672} 
& \third{18.7581} & \third{0.6651} & \third{0.3896} & \third{1032.4097}
& \third{21.9872} & \third{0.3301} \\

Gen3C  
& \third{0.9684} & \first{0.9695} & \first{0.9955} & \first{0.7250} & \first{0.5673} 
& \second{19.9375} & \second{0.6755} & \second{0.3630} & \second{756.7273}
& \second{5.7102} 
& \second{0.7578} \\

\textbf{VideoWeave-I2V} 
& 0.9573 
& \second{0.9684} 
& \second{0.9936} 
& \second{0.7143} 
& \third{0.5483} 
& \first{{21.1826}} 
& \first{{0.7275}} 
& \first{{0.3109}} 
& \first{{565.2975}} 
& \first{5.3582} 
& \first{0.7788} \\
\bottomrule
\end{tabular}
}
\end{table*}

\begin{figure*}[t]
    \centering
    \includegraphics[width=\textwidth]{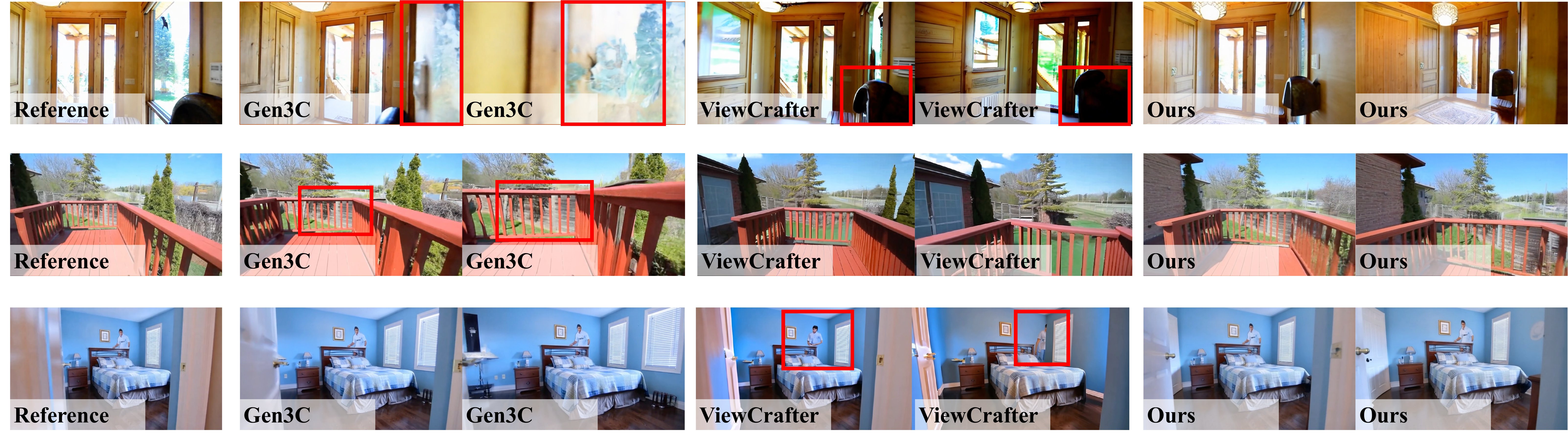}
\caption{
\textbf{Qualitative Comparison of Image-to-Video Generation.}
Methods that rely on explicit reconstruction results as generation guidance tend to propagate depth estimation errors into the video generator, leading to noticeable object penetration and spatial misalignment.
}
    \label{fig:i2v_qualitative_results}
\end{figure*}

\noindent \textbf{Image-to-Video Quantitative Results.}
As shown in Table~\ref{tab:main_results_i2v}, VideoWeave-I2V achieves the best performance on all 3D reconstruction and epipolar consistency metrics under the fixed reference image and camera trajectory setting.
GeometryForcing is not evaluated on epipolar metrics because its native output resolution is too low for reliable SIFT-based epipolar matching.
Compared with the larger-scale Gen3C, VideoWeave-I2V obtains comparable VBench scores, with close subject consistency, motion smoothness, imaging quality, and aesthetic quality, while further improving PSNR, SSIM, LPIPS, MSE, epipolar error, and inlier rate.
These results indicate that VideoWeave preserves the visual quality of strong I2V baselines while producing more geometrically consistent novel views.
This verifies the effectiveness of learning an implicit geometry--video prior aligned with the video generation process for camera-controllable video synthesis.

\begin{figure*}[t]
    \centering
    \includegraphics[width=\textwidth]{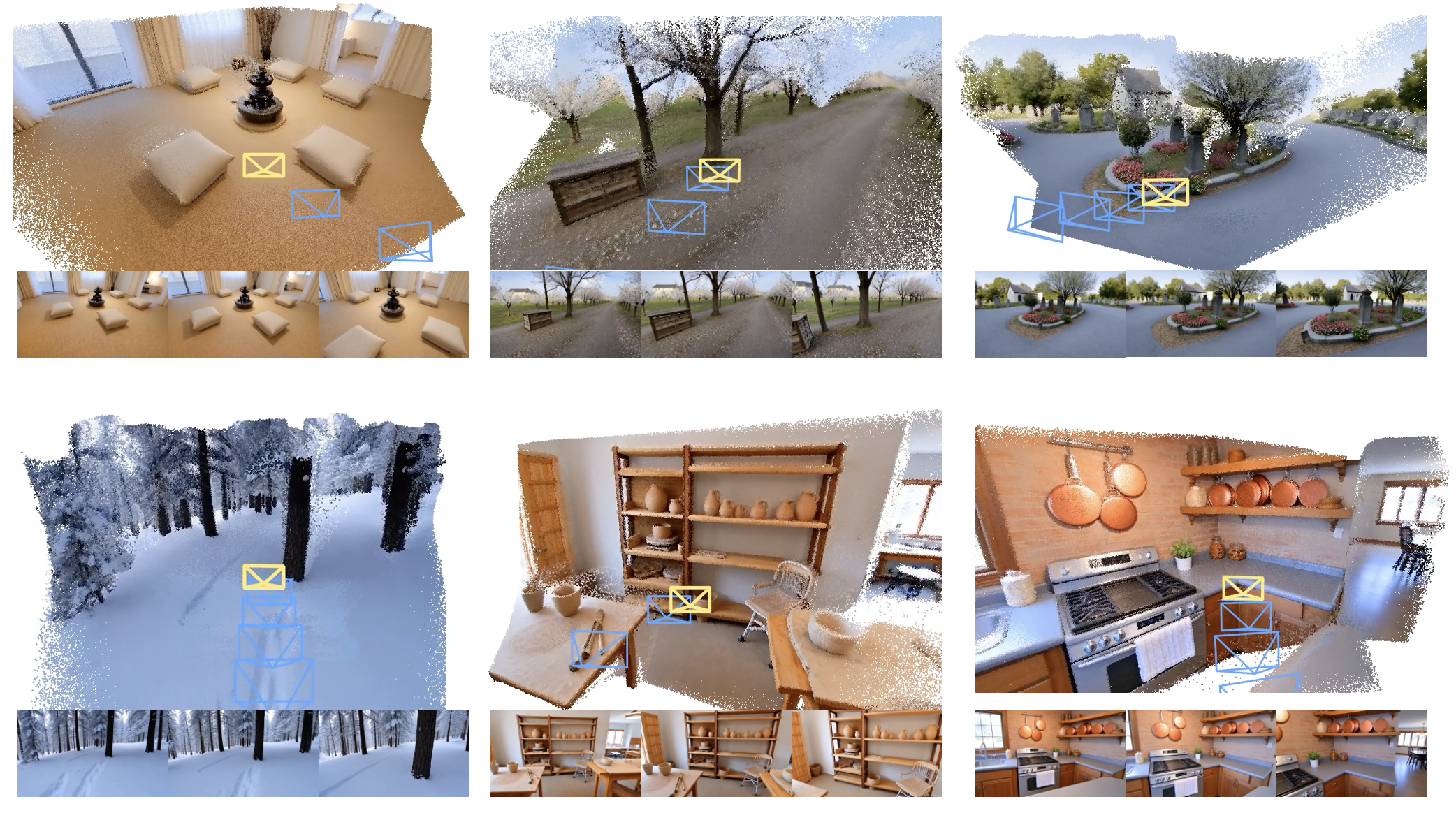}
    \caption{
    \textbf{Point Cloud Reconstruction from VideoWeave Videos.}
    We reconstruct 3D point clouds from videos generated by VideoWeave. 
    The results show coherent scene geometry, complete object structures, and consistent cross-view alignment under camera motion.
}
    \label{fig:pointcloud_results}
\end{figure*}

\begin{figure*}[t]
    \centering
    \includegraphics[width=\textwidth]{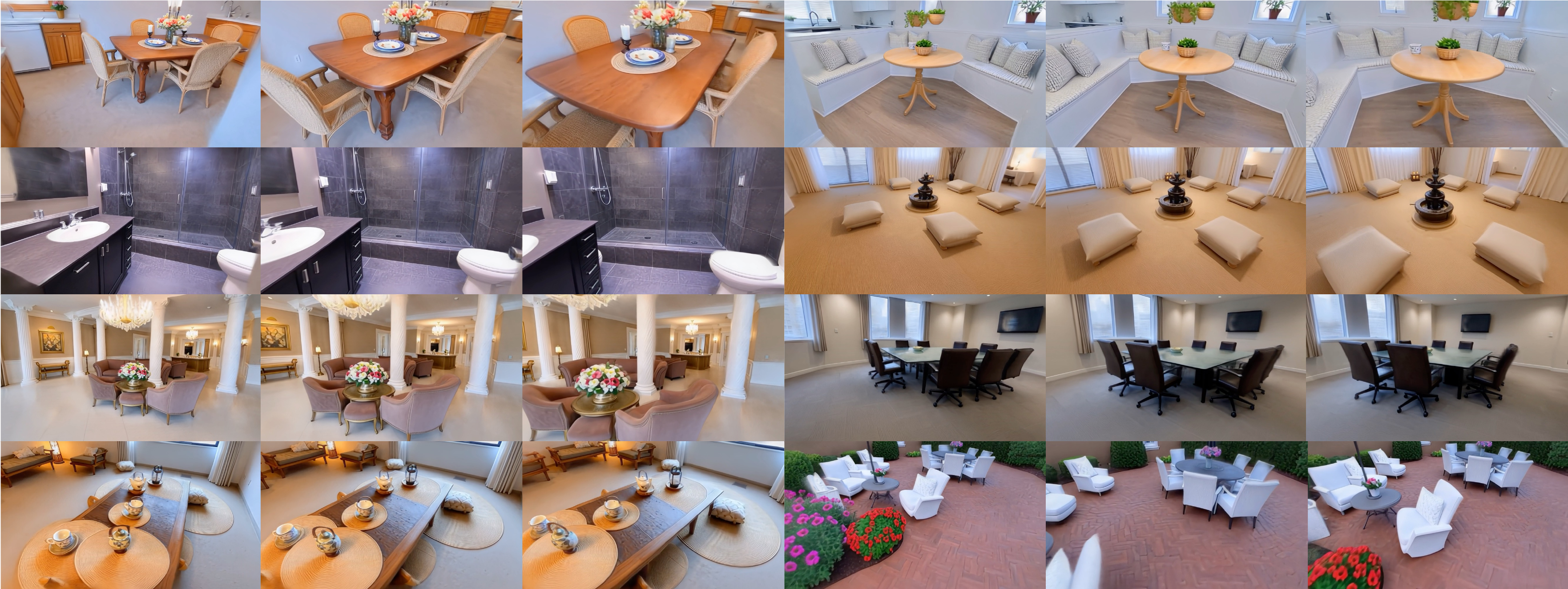}
    \caption{
    \textbf{3DGS Reconstruction from VideoWeave Videos.}
    We reconstruct videos generated by VideoWeave using 3D Gaussian Splatting and visualize the resulting 3D scenes. 
    The reconstructions preserve stable scene layouts and consistent geometry across viewpoints.
}
    \label{fig:3dgs_results}
\end{figure*}

\noindent \textbf{Image-to-Video Qualitative Results.}
We show qualitative I2V comparisons in Fig.~\ref{fig:i2v_qualitative_results}.
VideoWeave substantially improves geometric consistency under the same reference frame and target camera trajectory. Explicit geometry-guided methods rely on reconstructed results to guide generation. However, when the target camera largely deviates from the reference view, these reconstructed results become unreliable, often introducing missing regions, geometric stretching, object penetration, and spatial misalignment. Since such results are directly used to construct generation guidance, their errors are propagated into the generated video, leading to broken structures and inconsistent cross-view geometry. In contrast, VideoWeave does not decode depth, point clouds, or intermediate rendered views to build conditions, supervision targets, or rewards. It injects geometry through an implicit geometry--video joint prior aligned with the generation process, enabling more coherent spatial structure and more stable novel-view synthesis under large camera motion.

\noindent \textbf{More Qualitative Results.}
To further evaluate the geometric structure of the generated videos, we additionally provide 3D reconstruction results using point clouds and 3D Gaussian Splatting, as shown in Fig.~\ref{fig:pointcloud_results} and Fig.~\ref{fig:3dgs_results}.

\begin{table*}[t]
\centering
\caption{
    \textbf{Ablation on Feature Projection Strategy.}
    We compare naive PCA projection with the proposed learnable geometry adapter.
    The learnable adapter consistently improves video quality, reconstruction accuracy, and 3D consistency.
}
\label{tab:ablation_projection}
\vspace{-1mm}
\scriptsize
\setlength{\tabcolsep}{3.0pt}
\renewcommand{\arraystretch}{0.92}
\resizebox{0.98\textwidth}{!}{
\begin{tabular}{l | ccccc | cccc | cc}
\toprule
\multirow{2}{*}{\textbf{Method}} &
\multicolumn{5}{c|}{\textbf{VBench Metrics}} &
\multicolumn{4}{c|}{\textbf{3D Reconstruction Error}} &
\multicolumn{2}{c}{\textbf{3D Consistency Metrics}} \\
\cmidrule(lr){2-6} \cmidrule(lr){7-10} \cmidrule(l){11-12}
&
\makecell{\textbf{Background} \\ \textbf{Consistency} $\uparrow$} &
\makecell{\textbf{Subject} \\ \textbf{Consistency} $\uparrow$} &
\makecell{\textbf{Motion} \\ \textbf{Smoothness} $\uparrow$} &
\makecell{\textbf{Imaging} \\ \textbf{Quality} $\uparrow$} &
\makecell{\textbf{Aesthetic} \\ \textbf{Quality} $\uparrow$} &
\makecell{\textbf{PSNR} $\uparrow$} &
\makecell{\textbf{SSIM} $\uparrow$} &
\makecell{\textbf{LPIPS} $\downarrow$} &
\makecell{\textbf{MSE} $\downarrow$} &
\makecell{\textbf{Epipolar} \\ \textbf{Error} $\downarrow$} &
\makecell{\textbf{Inlier} \\ \textbf{Rate} $\uparrow$} \\
\midrule
PCA Projection    
& \second{0.9086} 
& \second{0.8240} 
& \second{0.9652} 
& \second{0.4305} 
& \second{0.4095} 
& \second{16.7129} 
& \second{0.5785} 
& \second{0.5043} 
& \second{1715.3236} 
& \second{29.9325} 
& \second{0.2216} \\

Learnable Adapter 
& \first{0.9305}
& \first{0.9530} 
& \first{0.9910} 
& \first{0.7383} 
& \first{0.5988}
& \first{21.4994} 
& \first{0.7082} 
& \first{0.2987} 
& \first{500.8694} 
& \first{7.2780} 
& \first{0.7068} \\

\bottomrule
\end{tabular}
}
\vspace{-2mm}
\end{table*}

\begin{table*}[t]
\centering
\caption{
    \textbf{Component Ablation.}
    We ablate two key components of VideoWeave: geometry-video joint modeling and joint-prior distillation.
    Removing either component degrades video quality and temporal geometric consistency, while the full model achieves the best overall performance.
}
\label{tab:ablation_components}
\vspace{-1mm}
\scriptsize
\setlength{\tabcolsep}{3.0pt}
\renewcommand{\arraystretch}{0.92}
\resizebox{0.98\textwidth}{!}{
\begin{tabular}{cc | ccccc | cccc | cc}
\toprule
\multirow{2}{*}{\makecell{\textbf{Joint} \\ \textbf{Modeling}}} &
\multirow{2}{*}{\makecell{\textbf{Prior} \\ \textbf{Distill.}}} &
\multicolumn{5}{c|}{\textbf{VBench Metrics}} &
\multicolumn{4}{c|}{\textbf{3D Reconstruction Error}} &
\multicolumn{2}{c}{\textbf{3D Consistency Metrics}} \\
\cmidrule(lr){3-7} \cmidrule(lr){8-11} \cmidrule(l){12-13}
& &
\makecell{\textbf{Background} \\ \textbf{Consistency} $\uparrow$} &
\makecell{\textbf{Subject} \\ \textbf{Consistency} $\uparrow$} &
\makecell{\textbf{Motion} \\ \textbf{Smoothness} $\uparrow$} &
\makecell{\textbf{Imaging} \\ \textbf{Quality} $\uparrow$} &
\makecell{\textbf{Aesthetic} \\ \textbf{Quality} $\uparrow$} &
\makecell{\textbf{PSNR} $\uparrow$} &
\makecell{\textbf{SSIM} $\uparrow$} &
\makecell{\textbf{LPIPS} $\downarrow$} &
\makecell{\textbf{MSE} $\downarrow$} &
\makecell{\textbf{Epipolar} \\ \textbf{Error} $\downarrow$} &
\makecell{\textbf{Inlier} \\ \textbf{Rate} $\uparrow$} \\
\midrule

$\times$ & $\times$
& 0.9086 & \third{0.9000} & 0.9712 & 0.6659 & 0.5633 
& \second{19.7444} & \second{0.5913} & \second{0.3790} & \second{815.3085} 
& 19.9983 & 0.3935 \\

$\times$ & \checkmark
& \third{0.9209}
& 0.8931
& \third{0.9718}
& \third{0.7160}
& \second{0.6103}
& 17.2730
& 0.5199
& 0.4414
& 1525.9563
& \third{18.2811}
& \third{0.4110} \\

\checkmark & $\times$
& \first{0.9317}
& \second{0.9488}
& \second{0.9891}
& \second{0.7234}
& \first{0.6189}
& \third{19.1706}
& \third{0.5728}
& \third{0.3841}
& \third{928.8936}
& \second{9.4925}
& \second{0.6737} \\

\checkmark & \checkmark
& \second{0.9305}
& \first{0.9530}
& \first{0.9910}
& \first{0.7383}
& \third{0.5988}
& \first{21.4994}
& \first{0.7082}
& \first{0.2987}
& \first{500.8694}
& \first{7.2780}
& \first{0.7068} \\

\bottomrule
\end{tabular}
}
\vspace{-2mm}
\end{table*}

\begin{figure}[t]
    \centering
    \includegraphics[width=\linewidth]{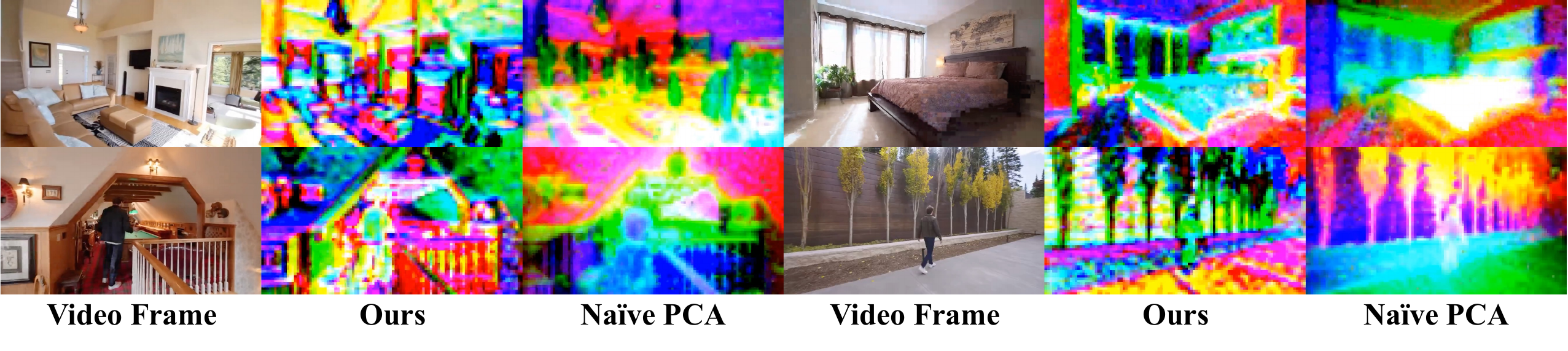}
    \caption{Visualization of projected geometry features using PCA and the learnable adapter.}
    \label{fig:pca}
\end{figure}

\begin{table*}[t]
\centering

\begin{minipage}[t]{0.485\textwidth}
\centering
\captionof{table}{
    \textbf{Ablation on Geometry Feature Dimension.}
    We vary the channel dimension $C_g$ of the adapted geometry latent while keeping the remaining pipeline unchanged.
}
\label{tab:ablation_geometry_dim}
\vspace{-1mm}
\scriptsize
\setlength{\tabcolsep}{3.0pt}
\renewcommand{\arraystretch}{0.92}
\resizebox{\linewidth}{!}{
\begin{tabular}{l | cccc | cc}
\toprule
\multirow{2}{*}{\textbf{Geometry Dimension}} &
\multicolumn{4}{c|}{\textbf{3D Reconstruction Error}} &
\multicolumn{2}{c}{\textbf{3D Consistency Metrics}} \\
\cmidrule(lr){2-5} \cmidrule(l){6-7}
&
\textbf{PSNR} $\uparrow$ &
\textbf{SSIM} $\uparrow$ &
\textbf{LPIPS} $\downarrow$ &
\textbf{MSE} $\downarrow$ &
\makecell{\textbf{Epipolar} \\ \textbf{Error} $\downarrow$} &
\makecell{\textbf{Inlier} \\ \textbf{Rate} $\uparrow$} \\
\midrule
\rowcolor{gray!15}
$C_g=4$
& \textbf{21.18} & \textbf{0.728} & \textbf{0.311} & \textbf{565.30} & 5.35 & 0.778 \\

$C_g=8$
& 20.41 & 0.710 & 0.323 & 677.20 & 6.15 & 0.755 \\

$C_g=16$
& 20.19 & 0.703 & 0.350 & 725.90 & \textbf{4.37} & \textbf{0.815} \\
\bottomrule
\end{tabular}
}
\end{minipage}
\hfill
\begin{minipage}[t]{0.485\textwidth}
\centering
\captionof{table}{
    \textbf{Ablation on the Joint Modeling Layer.}
    We compare different layers for intermediate geometry prediction under the same I2V training setting.
}
\label{tab:ablation_layer}
\vspace{-1mm}
\scriptsize
\setlength{\tabcolsep}{3.0pt}
\renewcommand{\arraystretch}{0.92}
\resizebox{\linewidth}{!}{
\begin{tabular}{l | cccc | cc}
\toprule
\multirow{2}{*}{\textbf{Geometry Prediction Layer}} &
\multicolumn{4}{c|}{\textbf{3D Reconstruction Error}} &
\multicolumn{2}{c}{\textbf{3D Consistency Metrics}} \\
\cmidrule(lr){2-5} \cmidrule(l){6-7}
&
\textbf{PSNR} $\uparrow$ &
\textbf{SSIM} $\uparrow$ &
\textbf{LPIPS} $\downarrow$ &
\textbf{MSE} $\downarrow$ &
\makecell{\textbf{Epipolar} \\ \textbf{Error} $\downarrow$} &
\makecell{\textbf{Inlier} \\ \textbf{Rate} $\uparrow$} \\
\midrule
Early Layer
& 20.62 & 0.706 & 0.325 & 648.98 & \textbf{5.20} & \textbf{0.783} \\

\rowcolor{gray!15}
Intermediate Layer
& \textbf{21.18} & \textbf{0.728} & \textbf{0.311} & \textbf{565.30} & 5.35 & 0.778 \\

Final Layer
& 20.80 & 0.725 & 0.322 & 621.01 & 5.55 & 0.768 \\
\bottomrule
\end{tabular}
}
\end{minipage}
\vspace{-3mm}
\end{table*}

\subsection{Ablations and Analysis}
We organize our ablations into component validation and design analysis.
For Factors 1--3, we evaluate the contribution of the core components of VideoWeave under the target text-to-video setting.
For Factors 4--6, we conduct the design ablations under an image-to-video setting because it provides a more controlled evaluation protocol.
In this setting, all variants share the same reference image and target camera trajectory, which reduces variations in generated content and motion and makes geometry-related differences easier to attribute to the design being examined.

\subsubsection{Component Ablation}

\noindent\textbf{Factor 1: How does feature projection affect geometry latent adaptation?}
We first analyze the effect of different geometry feature projection strategies. The features produced by the geometry estimator have a different distribution from the latent space of the video diffusion model. Therefore, directly applying PCA for linear projection may discard useful structural information. To verify this, we replace the learnable adapter with PCA projection. As shown in Table~\ref{tab:ablation_projection}, the PCA variant performs worse than the full model in both visual quality and geometric consistency, indicating that a simple linear projection is insufficient for converting geometry-model features into generation-compatible latents.
We further visualize the geometry features after the two projection strategies as shown in Figure~\ref{fig:pca}. The PCA features contain noticeable unclear local structure boundaries, and unstable fine-grained details. In contrast, the features adapted by the learnable adapter show clearer structural regions, and preserve more recognizable local geometric patterns. These observations suggest that the learnable adapter does more than reduce feature dimensionality. It also helps preserve geometry information relevant to video generation, leading to better visual quality and geometric consistency.

\noindent\textbf{Factor 2: How does joint geometry-video modeling affect 3D consistency?}
We evaluate standard video-only post-training and our geometry-video joint post-training.
In Table~\ref{tab:ablation_components}, the variant without joint modeling or distillation corresponds to the same-budget Wan-SFT baseline; the variant with distillation only uses video-only prior distillation; and the variant with joint modeling only evaluates the joint-trained generator before score distillation.
As shown in Table~\ref{tab:ablation_components}, removing joint modeling causes a substantial drop in 3D reconstruction quality and epipolar consistency.
This indicates that video-only post-training cannot reliably capture the coupling between visual motion and underlying geometry.
By modeling geometry and video as a unified latent distribution, our method transfers a stronger geometry-aware prior to the video generator, leading to better multi-view consistency, and preserved general video quality.

\noindent\textbf{Factor 3: How does joint score distillation affect the final generator?}
We further ablate the joint score distillation stage by training the model with geometry-video joint supervision only.
As shown in Table~\ref{tab:ablation_components}, removing distillation keeps competitive VBench scores on some appearance-related metrics, but clearly degrades 3D reconstruction quality and epipolar consistency.
This indicates that joint-modal post-training alone improves geometry-aware representations, but is insufficient to fully transfer the geometry-video joint prior into the student generator.
With joint score distillation, the student is supervised by the geometry-video joint teacher at the distribution level, rather than relying only on direct feature regression or decoded reconstruction targets.
This enables the student to inherit 3D-aware generation behavior from the teacher and better coordinate visual dynamics with implicit geometry.
As a result, the full model produces stronger reconstruction quality, more stable cross-frame structures, and better epipolar consistency, while preserving competitive visual quality.

\subsubsection{Design Ablation}

\noindent\textbf{Factor 4: How does the geometry feature dimension affect joint modeling?}
We vary the channel dimension $C_g$ of the geometry latent after the adapter projection in Eq.~\ref{eq:geometry_adapter}.
As shown in Table~\ref{tab:ablation_geometry_dim}, a small geometry feature dimension is already sufficient for effective joint modeling.
The $C_g=4$ setting achieves the best reconstruction quality across PSNR, SSIM, LPIPS, and MSE, while maintaining strong epipolar consistency.
Increasing the dimension does not lead to monotonic improvements: $C_g=16$ obtains the lowest epipolar error and highest inlier rate, but its reconstruction metrics drop compared with $C_g=4$.
This suggests that additional geometry channels can strengthen sparse correspondence-level consistency, but also increase the burden of jointly modeling geometry and video appearance under limited model capacity.
Therefore, we adopt a small geometry feature dimension, which provides the best overall balance between reconstruction fidelity and geometric consistency.

\noindent\textbf{Factor 5: How does the joint modeling layer affect geometry--video learning?}
We ablate the network layer $\ell$ where the geometry prediction loss in Eq.~\ref{eq:joint_diffusion_loss} is applied.
As shown in Table~\ref{tab:ablation_layer}, applying geometry prediction at an intermediate layer achieves the best reconstruction quality, with the highest PSNR and SSIM and the lowest LPIPS and MSE.
Although the early-layer variant obtains slightly better epipolar error and inlier rate, the intermediate layer provides the best overall trade-off between reconstruction fidelity and geometric consistency.
This result indicates that geometry supervision is not necessarily more effective when applied closer to the output.
Early layers mainly capture local textures and low-level visual patterns, and therefore do not yet provide stable spatio-temporal layout representations.
In contrast, late layers are already highly coupled with the final synthesis of video latents.
Imposing geometry supervision at these layers can interfere with the pretrained video prior and introduce a conflict between appearance generation and geometry prediction.
Intermediate layers provide a more suitable interface for geometry--video interaction: they encode relatively stable scene structures and cross-frame relations, while remaining less directly tied to the final pixel-level generation pathway.
Therefore, geometry constraints can be introduced at this stage without substantially disrupting the original video generation capability, leading to a more balanced performance profile.

\begin{wraptable}{r}{0.49\textwidth}
\vspace{-4mm}
\centering
\caption{
    \textbf{Ablation on Geometry Representation Level.}
    We compare features from different levels of the geometry backbone.
}
\label{tab:ablation_geometry_feature_level}
\vspace{-1mm}
\scriptsize
\setlength{\tabcolsep}{3.0pt}
\renewcommand{\arraystretch}{0.92}
\resizebox{\linewidth}{!}{
\begin{tabular}{cc | cccc | cc}
\toprule
\multicolumn{2}{c|}{\textbf{Feature Level}} &
\multicolumn{4}{c|}{\textbf{3D Reconstruction Error}} &
\multicolumn{2}{c}{\textbf{3D Consistency Metrics}} \\
\cmidrule(lr){1-2} \cmidrule(lr){3-6} \cmidrule(l){7-8}
\textbf{Low} & \textbf{High} &
\textbf{PSNR} $\uparrow$ &
\textbf{SSIM} $\uparrow$ &
\textbf{LPIPS} $\downarrow$ &
\textbf{MSE} $\downarrow$ &
\makecell{\textbf{Epipolar} \\ \textbf{Error} $\downarrow$} &
\makecell{\textbf{Inlier} \\ \textbf{Rate} $\uparrow$} \\
\midrule
\checkmark & $\times$
& 20.53 & 0.706 & 0.335 & 649.99 & \textbf{4.34} & \textbf{0.823} \\

$\times$ & \checkmark
& 20.84 & 0.718 & 0.359 & 610.61 & 8.78 & 0.642 \\

\rowcolor{gray!15}
\checkmark & \checkmark
& \textbf{21.18} & \textbf{0.728} & \textbf{0.311} & \textbf{565.30} & 5.35 & 0.778 \\
\bottomrule
\end{tabular}
}
\vspace{-3mm}
\end{wraptable}

\noindent\textbf{Factor 6: How does the level of geometry representation affect joint modeling?}
We further study how geometry representations from different feature levels affect joint modeling.
As shown in Table~\ref{tab:ablation_geometry_feature_level}, using a single feature level provides only a partial geometry cue for joint modeling.
Low-level features emphasize local geometric details, while high-level features encode more global scene structure.
Combining both levels provides a richer geometry representation that captures fine-grained spatial correspondences and scene-level layout cues in a unified latent.
The low-level-only variant achieves the best epipolar error and inlier rate, indicating that local geometry features are effective for sparse cross-frame correspondence.
However, the combined representation achieves the best reconstruction quality across PSNR, SSIM, LPIPS, and MSE, while maintaining strong epipolar consistency.
Therefore, we use the combined multi-level geometry representation as the input to $\mathbf{g}_\psi$, since it provides a more balanced and transferable structural prior for geometry--video joint diffusion.

\section{Conclusion}
We introduced \textbf{VideoWeave}, a latent-space post-training framework that transfers 3D-aware priors from implicit geometry latents into pretrained video diffusion models. Instead of decoding these latents into explicit depth, point clouds, reconstructed 3D assets, or rendered views for conditions, supervision targets, or rewards, VideoWeave progressively adapts implicit geometry features, learns a joint geometry-video diffusion prior, and distills the resulting joint score field into a compact student generator. The distilled student preserves this geometry-aware prior without requiring a geometry encoder, reconstruction module, or renderer at inference time. Together with the curated \textbf{GeoVid-80K} dataset, this framework enables efficient geometry-aware video generation. Experiments across visual quality, 3D reconstruction, and epipolar consistency demonstrate that VideoWeave substantially improves spatial coherence and geometric stability while preserving the original generative fidelity. These results demonstrate that implicit geometry-video joint modeling provides an effective and scalable path toward world-consistent video generation.

\bibliography{2026_conference}
\bibliographystyle{2026_conference}


\end{document}